\pdfoutput=1

\documentclass[11pt]{article}

\usepackage{acl}

\usepackage{times}
\usepackage{latexsym}

\usepackage[T1]{fontenc}

\usepackage[utf8]{inputenc}

\usepackage{microtype}
\usepackage{color}
\usepackage{bm}
\usepackage{amssymb}
\usepackage{amsfonts}
\usepackage{amsmath}
\usepackage{footmisc}
\usepackage{multirow}
\usepackage{multicol}
\usepackage{caption}
\usepackage{subcaption}
\usepackage{latexsym}
\usepackage{mflogo}
\usepackage{amsthm}
\usepackage{graphicx} 
\usepackage{booktabs}
\usepackage{xspace}
\usepackage{newfloat}
\usepackage{listings}
\usepackage[ruled,linesnumbered]{algorithm2e}

\newcommand{\paratitle}[1]{\vspace{1.5ex}\noindent\textbf{#1}}
\newcommand{\ie}{\emph{i.e.,}\xspace}

\newcommand{\eg}{\emph{e.g.,}\xspace}

\newcommand{\ignore}[1]{}

\title{Debiased Contrastive Learning of Unsupervised Sentence Representations}

\author{
\setcounter{footnote}{1}
	Kun Zhou\textsuperscript{\rm{1},\rm{3}},
	Beichen Zhang\textsuperscript{\rm{2}},
	Wayne Xin Zhao\textsuperscript{\rm{2},\rm{3}}\thanks{$^\dagger$ Corresponding author} \and
	\textbf{Ji-Rong Wen}\textsuperscript{\rm{2},\rm{3}} \\
	\textsuperscript{1}School of Information, Renmin University of China. \\
	\textsuperscript{2}Gaoling School of Artificial Intelligence, Renmin University of China \\
	\textsuperscript{3}Beijing Key Laboratory of Big Data Management and Analysis Methods\\
	\texttt{francis\_kun\_zhou@163.com}, \texttt{\{zhangbeichen724,batmanfly\}@gmail.com},\\
	\texttt{jrwen@ruc.edu.cn} \\
}

\begin{document}
\maketitle
\begin{abstract}
Recently, contrastive learning has been shown to be effective in improving pre-trained language models (PLM) to derive high-quality sentence representations. It aims to pull close positive examples  to enhance the alignment while push apart irrelevant negatives for the uniformity of the whole representation space.
However, previous works mostly adopt in-batch negatives or sample from training data at random. 
Such a way may cause the sampling bias that improper negatives (\eg false negatives and anisotropy representations) are used to learn sentence representations, which will hurt the uniformity of the representation space.
To address it, we present a new framework \textbf{DCLR} (\underline{D}ebiased \underline{C}ontrastive \underline{L}earning of unsupervised sentence \underline{R}epresentations) to alleviate the influence of these improper negatives.
In DCLR, we design an instance weighting method to punish false negatives and generate noise-based negatives to guarantee the uniformity of the representation space.
Experiments on seven semantic textual similarity tasks show that our approach is more effective than competitive baselines. Our code and data are publicly available at the link: \textcolor{blue}{\url{https://github.com/RUCAIBox/DCLR}}.
\end{abstract}
 
\section{Introduction}
As a fundamental task in the natural language processing (NLP) field, unsupervised sentence representation learning~\cite{DBLP:conf/nips/KirosZSZUTF15,DBLP:conf/naacl/HillCK16} aims to derive high-quality sentence representations that can benefit various downstream tasks, especially for low-resourced domains or computationally expensive tasks, \eg zero-shot text semantic matching~\cite{qiao2016less},
large-scale semantic similarity comparison~\cite{DBLP:conf/semeval/AgirreBCCDGGLMM15}, and document retrieval~\cite{DBLP:conf/icml/LeM14}.

Recently,  pre-trained language models~(PLMs) \cite{DBLP:conf/naacl/DevlinCLT19} have become a widely-used semantic representation approach, achieving remarkable performance on various NLP tasks.
However, several studies have found that the native sentence representations derived by PLMs are not uniformly distributed with respect to directions, but instead occupy a \emph{narrow cone} in the vector space~\cite{DBLP:conf/emnlp/Ethayarajh19}, which largely limits their expressiveness.
To address this issue, contrastive learning~\cite{DBLP:conf/icml/ChenK0H20} has been adopted to refine PLM-derived sentence representations.
It pulls semantically-close neighbors together to improve the alignment, while pushes apart non-neighbors for the uniformity of the whole representation space.
In the learning process, both positive and negative examples are involved in contrast with the original sentence. 
For positive examples, previous works apply data augmentation strategies~\cite{DBLP:conf/acl/YanLWZWX20} on the original sentence to generate highly similar variations. While, negative examples are commonly sampled from the batch or training data (\eg in-batch negatives~\cite{DBLP:conf/emnlp/GaoYC21}) at random, due to the lack of ground-truth annotations for negatives.

\begin{figure}[t]
\centering
\includegraphics[width=0.46\textwidth]{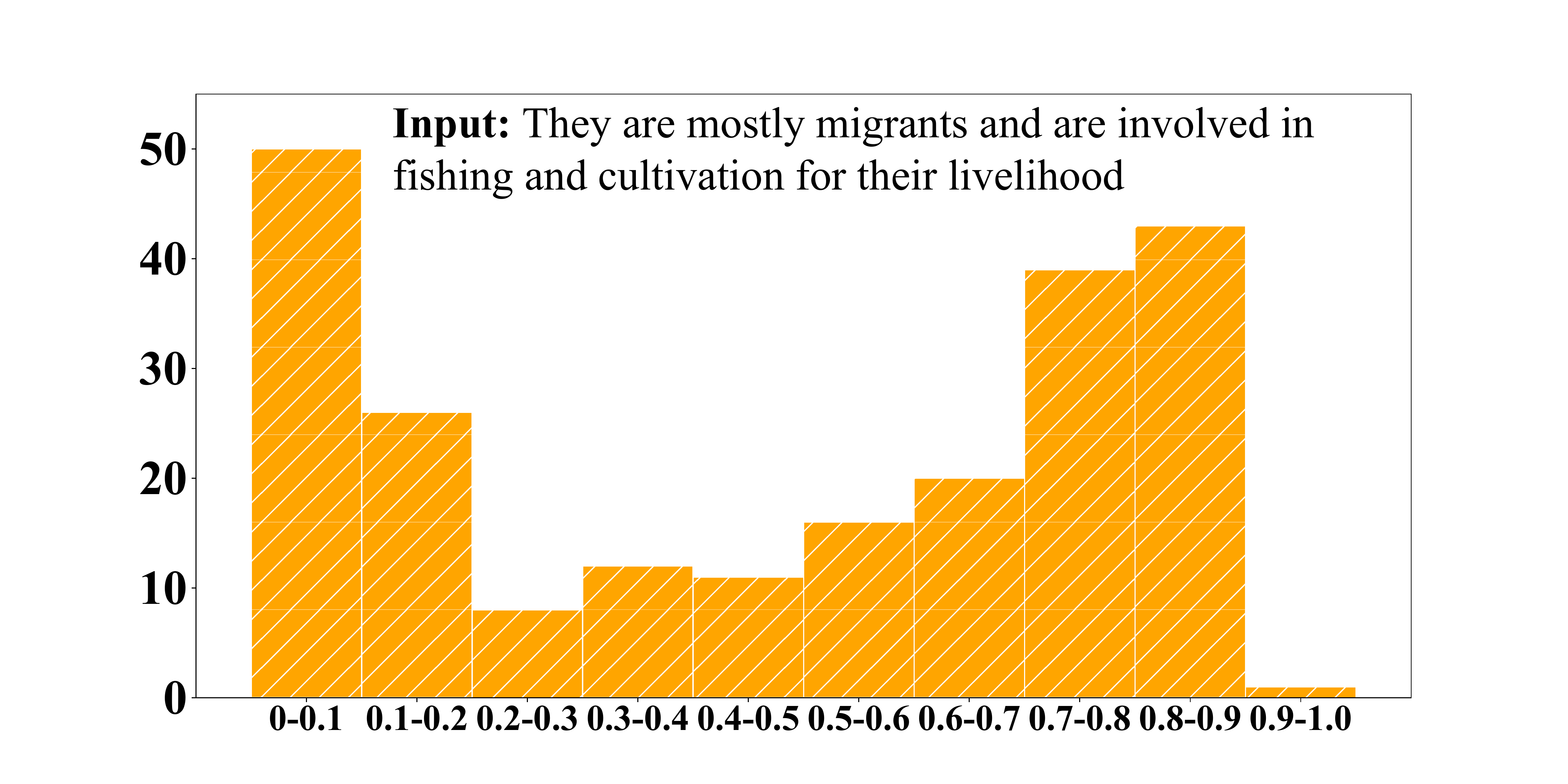}
\caption{The distribution of cosine similarity between an input sentence and 255 in-batch negatives from the commonly-used Wikipedia Corpus. It is evaluated by the SimCSE model~\cite{DBLP:conf/emnlp/GaoYC21}. Almost half of the negatives have high similarities with the input.}
\label{intro}
\end{figure}

Although such a negative sampling way is simple and convenient, it may cause \emph{sampling bias} and affect the sentence representation learning.
First, the sampled negatives are likely to be \emph{false negatives} that are indeed semantically close to the original sentence.
As shown in Figure~\ref{intro}, given an input sentence, about half of in-batch negatives have a cosine similarity above 0.7 with the original sentence based on the SimCSE model~\cite{DBLP:conf/emnlp/GaoYC21}.
It is likely to hurt the semantics of the sentence representations by simply pushing apart these sampled negatives.
Second, due to the anisotropy problem~\cite{DBLP:conf/emnlp/Ethayarajh19}, the representations of sampled negatives are from the narrow representation cone spanned by PLMs, which cannot fully reflect the overall semantics of the representation space. Hence, it is sub-optimal to only rely on these representations for learning the uniformity objective of sentence representations.

To address the above issues, we aim to develop a better contrastive learning approach with debiased negative sampling strategies.
The core idea is to improve the random negative sampling strategy for alleviating the sampling bias problem.
First, in our framework, we design an instance weighting method to punish the sampled false negatives during training. 
We incorporate a complementary model to evaluate the similarity between each negative and the original sentence, then assign lower weights for negatives with higher similarity scores.
In this way, we can detect semantically-close false negatives and further reduce their influence.
Second, we randomly initialize new negatives based on random Gaussian noises to simulate sampling within the whole semantic space, and devise a gradient-based algorithm to optimize the noise-based negatives towards the most nonuniform points.
By learning to contrast with the nonuniform noise-based negatives, we can extend the occupied space of sentence representations and improve the uniformity of the representation space.

To this end, we propose \textbf{DCLR}, a general framework towards \underline{D}ebiased \underline{C}ontrastive \underline{L}earning of unsupervised sentence \underline{R}epresentations.
In our approach, we first initialize the noise-based negatives from a Gaussian distribution, and leverage a gradient-based algorithm to update the new negatives by considering the uniformity of the representation space.
Then, we adopt the complementary model to produce the weights for these noise-based negatives and randomly sampled negatives, where the false negatives will be punished.
Finally, we augment the positive examples via dropout~\cite{DBLP:conf/emnlp/GaoYC21} and combine them with the above weighted negatives for contrastive learning.
We demonstrate that our DCLR outperforms a number of competitive baselines on seven semantic textual similarity (STS) tasks using BERT~\cite{DBLP:conf/naacl/DevlinCLT19} and RoBERTa~\cite{DBLP:journals/corr/abs-1907-11692}.

Our contributions are summarized as follows:

(1) To our knowledge, our approach is the first attempt to reduce the sampling bias in contrastive learning of unsupervised sentence representations.

(2) We propose DCLR, a debiased contrastive learning framework that incorporates an instance weighting method to punish false negatives and generates noise-based negatives to guarantee the uniformity of the representation space.

(3) Experimental results on seven semantic textual similarity tasks show the effectiveness of our framework.

\section{Related Work}
In this section, we review the related work from the following three aspects. 

\paratitle{Sentence Representation Learning.}
Learning universal sentence representations~\cite{DBLP:conf/nips/KirosZSZUTF15,DBLP:conf/naacl/HillCK16} is the key to the success of various downstream tasks.
Previous works can be roughly categorized into supervised~\cite{DBLP:conf/emnlp/ConneauKSBB17,DBLP:conf/emnlp/CerYKHLJCGYTSK18} and unsupervised approaches~\cite{DBLP:conf/naacl/HillCK16,DBLP:conf/emnlp/LiZHWYL20}.
Supervised approaches rely on annotated datasets (\eg NLI~\cite{DBLP:conf/emnlp/BowmanAPM15,DBLP:conf/naacl/WilliamsNB18}) to train the sentence encoder~\cite{DBLP:conf/emnlp/CerYKHLJCGYTSK18,DBLP:conf/emnlp/ReimersG19}.
Unsupervised approaches consider deriving sentence representations without labeled datasets, \eg 
pooling word2vec embeddings~\cite{DBLP:conf/nips/MikolovSCCD13}.
Recently, to leverage the strong potential of PLMs~\cite{DBLP:conf/naacl/DevlinCLT19}, several works propose to alleviate the anisotropy problem~\cite{DBLP:conf/emnlp/Ethayarajh19,DBLP:conf/emnlp/LiZHWYL20} of PLMs via special strategies, \eg flow-based approach~\cite{DBLP:conf/emnlp/LiZHWYL20} and whitening method~\cite{DBLP:journals/corr/abs-2104-01767}.
Besides, contrastive learning~\cite{DBLP:journals/corr/abs-2012-15466,DBLP:conf/emnlp/GaoYC21} has been used to refine the representations of PLMs.

\paratitle{Contrastive Learning.}
Contrastive learning has been originated in the computer vision~\cite{DBLP:conf/cvpr/HadsellCL06,DBLP:conf/cvpr/He0WXG20} and information retrieval~\cite{DBLP:conf/cikm/BianZZCHYW21,DBLP:conf/wsdm/ZhouZZWJ022} field with significant performance improvement. 
Usually, it relies on data augmentation strategies such as random cropping and image rotation~\cite{DBLP:conf/icml/ChenK0H20,DBLP:conf/acl/YanLWZWX20} to produce a set of semantically related positive examples for learning, and randomly samples negatives from the batch or whole dataset.
For sentence representation learning, contrastive learning can achieve a better balance between alignment and uniformity in semantic representation space.
Several works further adopt back translation~\cite{DBLP:journals/corr/abs-2005-12766}, token shuffling~\cite{DBLP:conf/acl/YanLWZWX20} and dropout~\cite{DBLP:conf/emnlp/GaoYC21} to augment positive examples for sentence representation learning.
However, the quality of the randomly sampled negatives is seldom studied.

\paratitle{Virtual Adversarial Training.}
Virtual adversarial training (VAT)~\cite{DBLP:journals/pami/MiyatoMKI19,DBLP:conf/iclr/KurakinGB17a} perturbs a given input with learnable noises to maximize the divergence of the model’s prediction with the original label, then utilizes the perturbed examples to improve the generalization~\cite{DBLP:conf/iclr/MiyatoDG17,DBLP:conf/iclr/MadryMSTV18}.
A class of VAT methods can be formulated into solving a min-max problem, which can be achieved by multiple projected gradient ascent steps~\cite{DBLP:conf/nips/QinMGKDFDSK19}.
In the NLP field, several studies incorporate adversarial perturbations in the embedding layer, and show its effectiveness on text classification~\cite{DBLP:conf/iclr/MiyatoDG17}, machine translation~\cite{DBLP:conf/coling/SunWCLUSZ20}, and natural language understanding~\cite{DBLP:conf/acl/JiangHCLGZ20} tasks.

\section{Preliminary}
This work aims to make use of unlabeled corpus for learning effective sentence representations that can be directly utilized for downstream tasks, \eg semantic textual similarity task~\cite{DBLP:conf/semeval/AgirreBCCDGGLMM15}.
Given a set of input sentences $\mathcal{X}=\{x_{1}, x_{2}, \dots, x_{n}\}$, our goal is to learn a representation ${h}_{i} \in \mathcal{R}^{d}$ for each sentence $x_{i}$ in an unsupervised manner.
For simplicity, we denote this process with a parameterized function ${h}_{i}=f(x_{i})$.

In this work, we mainly focus on using BERT-based PLMs~\cite{DBLP:conf/naacl/DevlinCLT19,DBLP:journals/corr/abs-1907-11692} to generate sentence representations.
Following existing works~\cite{DBLP:conf/emnlp/LiZHWYL20,DBLP:conf/acl/YanLWZWX20}, we fine-tune PLMs on the unlabeled corpus via our proposed unsupervised learning approach.
After that, for each sentence $x_{i}$, we encode it by the fine-tuned PLMs and take the representation of the \texttt{[CLS]} token from the last layer as its sentence representation ${h}_{i}$.

\section{Approach}
\begin{figure*}[t]
\centering
\includegraphics[width=\textwidth]{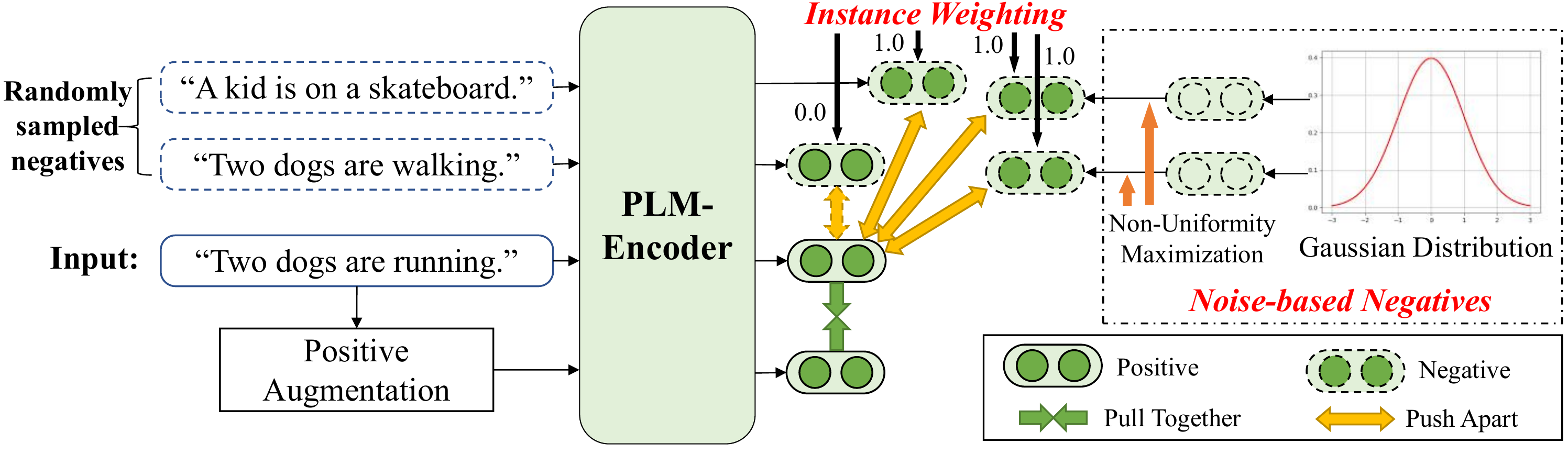}
\caption{The overview of our DCLR framework with noise-based negatives and the instance weighting strategy. We show the case that a false negative is punished by assigning the weight 0.}
\label{approach}
\end{figure*}

Our proposed framework DCLR focuses on reducing the influence of sampling bias in the contrastive learning of sentence representations.
In this framework, we devise a noise-based negatives generation strategy to reduce the bias caused by the anisotropy PLM-derived representations, and an instance weighting method to reduce the bias caused by false negatives.
Concretely, we initialize the noise-based negatives based on a Gaussian distribution and iteratively update these negatives towards non-uniformity maximization. 
Then, we utilize a complementary model to produce weights for all negatives (\ie randomly sampled and the noise-based ones).
Finally, we combine the weighted negatives and augmented positive examples for contrastive learning.
The overview of our DCLR is presented in Figure~\ref{approach}. 

\subsection{Generating Noise-based Negatives}
We aim to generate new negatives beyond the sentence representation space of PLMs during the training process, to alleviate the sampling bias derived from the anisotropy problem of PLMs~\cite{DBLP:conf/emnlp/Ethayarajh19}.
For each input sentence $x_{i}$, we first initialize $k$ noise vectors from a Gaussian distribution as the negative representations~\footnote{A contemporaneous work~\cite{DBLP:journals/corr/abs-2109-04321} has also utilized Gaussian distribution initialized noise vectors as smoothed representations for contrastive learning of sentence embeddings.}:
\begin{equation}
    \{\hat{h}_{1},\hat{h}_{2},\cdots,\hat{h}_{k}\} \sim \mathcal{N}(0, \sigma^{2}),
    \label{eq-gaussian}
\end{equation}
where $\sigma$ is the standard variance. 
Since these vectors are randomly initialized from such a Gaussian distribution, they are uniformly distributed within the whole semantic space.
By learning to contrast with these new negatives, it is beneficial for the uniformity of sentence representations.

To further improve the quality of the new negatives, we consider iteratively updating the negatives to capture the non-uniformity points within the whole semantic space.
Inspired by VAT~\cite{DBLP:conf/iclr/MiyatoDG17,DBLP:conf/iclr/ZhuCGSGL20}, we design a non-uniformity loss maximization objective to produce gradients for improving these negatives.
The non-uniformity loss is denoted as the contrastive loss between the noise-based negatives $\{\hat{h}_j\}$ and the positive representations of the original sentence $(h_{i}, h^{+}_{i})$ as:
\begin{equation}
\label{eq-cl}\small
L_{U}(h_{i}, h^{+}_{i}, \{\hat{h}\}) = -\log\frac{e^{\text{sim}(h_{i}, h^{+}_{i})/\tau_{u}}}{\sum_{\hat{h}_j\in\{\hat{h}_j\}}{e^{\text{sim}(h_{i},\hat{h}_i)/\tau_{u}}}},
\end{equation}
where $\tau_{u}$ is a temperature hyper-parameter and $\text{sim}(h_{i}, h^{+}_{i})$ is the cosine similarity $\frac{h_{i}^{\top}h^{+}_{i}}{||h_{i}||\cdot||h^{+}_{i}||}$. 
Based on it, for each negative $\hat{h}_j\in\{\hat{h}\}$, we optimize it by $t$ steps gradient ascent as
\begin{align}
\label{eq-pgd-t}
\hat{h}_j&=\hat{h}_j+\beta g(\hat{h}_j)/||g(\hat{h}_j)||_{2}, \\
g(\hat{h}_j)&=\bigtriangledown_{\hat{h}_j}L_{U}(h_{i}, h^{+}_{i}, \{\hat{h}\}),
\label{eq-pgd-f}
\end{align}
where $\beta$ is the learning rate, $||\cdot||_{2}$ is the $L_2$-norm. $g(\hat{h}_j)$ denotes the gradient of $\hat{h}_j$ by maximizing the non-uniformity loss between the positive representations and the noise-based negatives.
In this way, the noise-based negatives will be optimized towards the non-uniform points of the sentence representation space.
By learning to contrast with these negatives, the uniformity of the representation space can be further improved, which is essential for effective sentence representations.

\subsection{Contrastive Learning with Instance Weighting}
In addition to the above noise-based negatives, we also follow existing works~\cite{DBLP:conf/acl/YanLWZWX20,DBLP:conf/emnlp/GaoYC21} that adopt other in-batch representations as negatives $\{\Tilde{h}^{-}\}$.
However, as discussed before, the sampled negatives may contain examples that have similar semantics with the positive example (\ie false negatives).

To alleviate this problem, we propose an instance weighting method to punish the false negatives.
Since we cannot obtain the true labels or semantic similarities, we utilize a complementary model to produce the weights for each negative.
In this paper, we adopt the state-of-the-art SimCSE~\cite{DBLP:conf/emnlp/GaoYC21} as the complementary model.~\footnote{For convenience, we utilize SimCSE on BERT-base or RoBERTa-base model as the complementary model.}
Given a negative representation $h^{-}$ from $\{\Tilde{h}^{-}\}$ or $\{\hat{h}\}$ and the representation of the original sentence $h_i$, we utilize the complementary model to produce the weight as
\begin{equation}
\label{eq-gate}
    \alpha_{h^{-}} = \begin{cases} 0, \text{sim}_{C}(h_{i},h^{-}) \geq \phi \\ 1, \text{sim}_{C}(h_{i},h^{-}) < \phi \end{cases}
\end{equation}
where $\phi$ is a hyper-parameter of the instance weighting threshold, and $\text{sim}_{C}(h_{i},h^{-})$ is the cosine similarity score evaluated by the complementary model. In this way, the negative that has a higher semantic similarity with the representation of the original sentence will be regarded as a false negative and will be punished by assigning the weight 0.
Based on the weights, we optimize the sentence representations with a debiased cross-entropy contrastive learning loss function as
\begin{equation}
\label{eq-cl}
L = -\log\frac{e^{\text{sim}(h_{i}, h^{+}_{i})/\tau}}{\sum_{h^{-}\in\{\hat{h}\}\cup\{\Tilde{h}^{-}\} }{\alpha_{h^{-}}\times e^{\text{sim}(h_{i},h^{-})/\tau}}},
\end{equation}
where $\tau$ is a temperature hyper-parameter.
In our framework, we follow SimCSE~\cite{DBLP:conf/emnlp/GaoYC21} that utilizes dropout to augment positive examples $h^{+}_{i}$.
Actually, we can utilize various positive augmentation strategies, and will investigate it in Section~\ref{sec:others}.

\subsection{Overview and Discussion}
In this part, we present the overview and discussion of our DCLR approach.

\subsubsection{Overview of DCLR}
Our framework DCLR contains three major steps. 
In the first step, we generate noise-based negatives to extend  in-batch negatives.
Concretely, we first initialize a set of new negatives via random Gaussian noises using Eq.~\ref{eq-gaussian}.
Then, we incorporate a gradient-based algorithm to adjust the noise-based negatives by maximizing the non-uniform objective using Eq.~\ref{eq-pgd-t}.
After several iterations, we can obtain the noise-based negatives that correspond to the nonuniform points within the whole semantic space, and we mix up them with in-batch negatives to compose the negative set.
In the second step, we adopt a complementary model (\ie SimCSE) to compute the semantic similarity between the original sentence and each example from the negative set, and produce the weights using Eq.~\ref{eq-gate}.
Finally, we augment the positive examples via dropout and utilize the negatives with corresponding weights for contrastive learning using Eq.~\ref{eq-cl}.

\subsubsection{Discussion}
As mentioned above, our approach aims to reduce the influence of the \emph{sampling bias} about the negatives, and is agnostic to various positive data augmentation methods (\eg token cutoff and dropout).
Compared with traditional contrastive learning methods~\cite{DBLP:conf/acl/YanLWZWX20,DBLP:conf/emnlp/GaoYC21}, our proposed DCLR expands the negative set by introducing noise-based negatives $\{\hat{h}\}$, and adds a weight term $\alpha_{h^{-}}$ to punish false negatives.
Since the noise-based negatives are initialized from a Gaussian distribution and do not correspond to real sentences, they are highly confident negatives to broaden the representation space.
By learning to contrast with them, the learning of the contrastive objective will not be limited by the anisotropy representations derived from PLMs.
As a result, the sentence representations can span a broader semantic space, and the uniformity of the representation semantic space can be improved.

Besides, our instance weighting method also alleviates the false negative problem caused by the randomly sampling strategy.
With the help of a complementary model, the false negatives with similar semantics as the original sentence will be detected and punished.

\section{Experiment - Main Results}
\begin{table*}[t!]
\begin{center}
\centering
\small
\begin{tabular}{l|l|ccccccc|c}
\hline
& \textbf{Models} & \textbf{STS12} & \textbf{STS13} & \textbf{STS14} & \textbf{STS15} & \textbf{STS16} & \textbf{STS-B} & \textbf{SICK-R} & \textbf{Avg.} \\
\hline
\hline
\multirow{2} * {\textbf{Non-BERT}} & GloVe (avg.)$^\dagger$ & 55.14 & 70.66 & 59.73 & 68.25 & 63.66 & 58.02 & 53.76 & 61.32 \\
& USE$^\dagger$ & 64.49 & 67.80 & 64.61 & 76.83 & 73.18 & 74.92 & 76.69 & 71.22 \\
\hline
\multirow{7} * {\textbf{BERT-base}} & CLS$^\dagger$ & 21.54 & 32.11 & 21.28 & 37.89 & 44.24 & 20.30 & 42.42 & 31.40 \\
& Mean$^\dagger$ & 30.87 & 59.89 & 47.73 & 60.29 & 63.73 & 47.29 & 58.22 & 52.57 \\
&First-Last AVG$^\ddagger$. & 39.70&	59.38&	49.67&	66.03&	66.19&	53.87&	62.06&	56.70\\ 
&+flow$^\ddagger$ & 58.40&	67.10&	60.85&	75.16&	71.22&	68.66&	64.47&	66.55 \\ 
&+whitening$^\ddagger$ & 57.83& 66.90 & 60.90 & 75.08& 71.31& 68.24& 63.73& 66.28\\ 
& +Contrastive~(BT)$^\dagger$ &54.26 &64.03 &54.28 &68.19 &67.50 &63.27 &66.91 &62.63\\
&+ConSERT & 64.64 & 78.49 & 69.07 & 79.72 & 75.95 & 73.97 & 67.31 & 72.74 \\
&+SG-OPT$^\dagger$ & 66.84 & 80.13 & 71.23 & 81.56 & 77.17 & 77.23 & 68.16 & 74.62\\
&+SimCSE  & \underline{68.40}&	\underline{82.41} &	\underline{74.38}&	\underline{80.91}&	\textbf{78.56}&	\underline{76.85}&	\textbf{72.23}&	\underline{76.25}\\
&+DCLR (Ours) & \textbf{70.81} & \textbf{83.73} & \textbf{75.11} & \textbf{82.56} & \underline{78.44} & \textbf{78.31} & \underline{71.59} & \textbf{77.22}\\
\hline
\multirow{7} * {\textbf{BERT-large}} & CLS$^\dagger$ & 27.44 & 30.76 & 22.59 & 29.98 & 42.74 & 26.75 & 43.44 & 31.96\\
& Mean$^\dagger$ & 27.67 & 55.79 & 44.49 & 51.67 & 61.88 & 47.00 & 53.85 & 48.91 \\
& First-Last AVG & 57.73 &61.17 &61.18 &68.07 &70.25 &59.59 &60.34 & 62.62\\
& +flow$^\dagger$ & 62.82 & 71.24 & 65.39 & 78.98 & 73.23 & 72.72 & 63.77 & 70.07 \\
& +whitening & 64.34 & 74.60 & 69.64 & 74.68 & 75.90 & 72.48 & 60.80 & 70.35\\ 
& +Contrastive~(BT)$^\dagger$ &52.04 &62.59 &54.25 &71.07 &66.71 &63.84 &66.53  &62.43\\
& +ConSERT & 70.69 & 82.96 & 74.13 & 82.78 & 76.66 & 77.53 & 70.37 & 76.45 \\
& +SG-OPT$^\dagger$ & 67.02 &79.42 &70.38 &81.72 &76.35 &76.16 &70.20 &74.46\\
& +SimCSE & \underline{70.88} & \underline{84.16} & \underline{76.43} & \underline{84.50} & \underline{79.76} & \underline{79.26} & \underline{73.88} & \underline{78.41} \\
& +DCLR (Ours) & \textbf{71.87} & \textbf{84.83} & \textbf{77.37} & \textbf{84.70} & \textbf{79.81} & \textbf{79.55} & \textbf{74.19} & \textbf{78.90}\\
\hline
\multirow{7} * {\textbf{RoBERTa-base}} & CLS$^\dagger$ & 16.67 & 45.57 & 30.36 & 55.08 & 56.98 & 45.41 & 61.89 & 44.57\\
& Mean$^\dagger$ & 32.11 &56.33 &45.22 &61.34 &61.98 &54.53 &62.03 &53.36\\
& First-Last AVG$^\ddagger$ & 40.88&	58.74&	49.07&	65.63&	61.48&	58.55&	61.63&	56.57\\
& +whitening$^\ddagger$ & 46.99 & 63.24 &	57.23 &	71.36 &	68.99 &	61.36 &	62.91 & 61.73\\ 
& +Contrastive~(BT)$^\dagger$ &62.34 &78.60 &68.65 &79.31 &77.49 &79.93 &71.97 &74.04 \\
& +SG-OPT$^\dagger$ & 62.57 & 78.96 & 69.24 & 79.99 & 77.17 & 77.60 & 68.42 &73.42\\
& +SimCSE & \textbf{70.16}&	\underline{81.77}&	\underline{73.24}&	\underline{81.36}&	\underline{80.65}&	\underline{80.22}&	\underline{68.56}&	\underline{76.57}\\
& +DCLR (Ours) & \underline{70.01} & \textbf{83.08} & \textbf{75.09} & \textbf{83.66} & \textbf{81.06} & \textbf{81.86} & \textbf{70.33} & \textbf{77.87} \\
\hline
\multirow{7} * {\textbf{RoBERTa-large}} & CLS$^\dagger$ &19.25 &22.97 &14.93 &33.41 &38.01 &12.52 &40.63 &25.96\\
& Mean$^\dagger$ &33.63 &57.22 &45.67 &63.00 &61.18 &47.07 &58.38 &52.31\\
& First-Last AVG & 58.91 & 58.62 & 61.44 & 69.05 & 65.23 & 59.38 & 58.84 & 61.64 \\
& +whitening & 64.17 & 73.92 & 71.06 & 76.40 & 74.87 & 71.68 & 58.49 & 70.08 \\
& +Contrastive~(BT)$^\dagger$ & 57.60 & 72.14 & 62.25 & 71.49 & 71.75 & 77.05 & 67.83 & 68.59\\
& +SG-OPT$^\dagger$ & 64.29 & 76.36 & 68.48 & 80.10 & 76.60 & 78.14 & 67.97 & 73.13\\
& +SimCSE & \underline{72.86} &	\underline{83.99} &	\underline{75.62} & \underline{84.77} & \underline{81.80} & \underline{81.98} &	\underline{71.26}&	\underline{78.90} \\
& +DCLR (Ours) & \textbf{73.09} & \textbf{84.57} & \textbf{76.13} & \textbf{85.15} & \textbf{81.99} & \textbf{82.35} & \textbf{71.80} & \textbf{79.30}\\
\hline
\hline
\end{tabular}
\end{center}

\caption{
Sentence embedding performance on STS tasks (Spearman's correlation).
The best performance and the second-best performance methods are denoted in bold and underlined fonts respectively.
$\dagger$: results from \citet{DBLP:conf/acl/KimYL20};
$\ddagger$: results from \citet{DBLP:conf/emnlp/GaoYC21};
all other results are reproduced or reevaluated by ourselves.
}
\label{tab:main_sts}
\end{table*}

\subsection{Experiment Setup}
Following previous works~\cite{DBLP:conf/acl/KimYL20,DBLP:conf/emnlp/GaoYC21}, we conduct experiments on seven standard STS tasks.
For all these tasks, we use the SentEval toolkit~\cite{DBLP:conf/lrec/ConneauK18} for evaluation.

\paratitle{Semantic Textual Similarity Task.}
We evaluate our approach on 7 STS tasks: STS 2012–2016~\cite{DBLP:conf/semeval/AgirreCDG12,DBLP:conf/starsem/AgirreCDGG13,DBLP:conf/semeval/AgirreBCCDGGMRW14,DBLP:conf/semeval/AgirreBCCDGGLMM15,DBLP:conf/semeval/AgirreBCDGMRW16}, STS Benchmark~\cite{DBLP:conf/semeval/CerDALS17} and SICK-Relatedness~\cite{DBLP:conf/lrec/MarelliMBBBZ14}.
These datasets contain pairs of two sentences, whose similarity scores are labeled from 0 to 5. The relevance between gold annotations and the scores predicted by sentence representations is measured by the Spearman correlation.
Following the suggestions from previous works~\cite{DBLP:conf/emnlp/GaoYC21,DBLP:conf/emnlp/ReimersG19}, we directly compute the cosine similarity between sentence embeddings for all STS tasks.

\paratitle{Baseline Methods.}
We compare DCLR with competitive unsupervised sentence representation learning methods, consisting of non-BERT and BERT-based methods:

(1) \textbf{GloVe}~\cite{DBLP:conf/emnlp/PenningtonSM14} averages GloVe embeddings of words as the sentence representation.

(2) \textbf{USE}~\cite{DBLP:conf/emnlp/CerYKHLJCGYTSK18} utilizes a Transformer model that learns the objective of reconstructing the surrounding sentences within a passage.

(3) \textbf{CLS}, \textbf{Mean} and \textbf{First-Last AVG}~\cite{DBLP:conf/naacl/DevlinCLT19} adopt the \texttt{[CLS]} embedding, mean pooling of token representations, average representations of the first and last layers as sentence representations, respectively.

(4) \textbf{Flow}~\cite{DBLP:conf/emnlp/LiZHWYL20} applies mean pooling on the layer representations and maps the outputs to the Gaussian space as sentence representations.

(5) \textbf{Whitening}~\cite{DBLP:journals/corr/abs-2103-15316} uses the whitening operation to refine representations and reduce dimensionality.

(6) \textbf{Contrastive (BT)}~\cite{DBLP:journals/corr/abs-2005-12766} uses contrastive learning with back-translation for data augmentation to enhance sentence representations.

(7) \textbf{ConSERT}~\cite{DBLP:conf/acl/YanLWZWX20} explores various text augmentation strategies for contrastive learning of sentence representations.

(8) \textbf{SG-OPT}~\cite{DBLP:conf/acl/KimYL20} proposes a contrastive learning method with a self-guidance mechanism for improving the sentence embeddings of PLMs.

(9) \textbf{SimCSE}~\cite{DBLP:conf/emnlp/GaoYC21} proposes a simple contrastive learning framework that utilizes dropout for data augmentation.

\paratitle{Implementation Details.}
We implement our model based on Huggingface's transformers~\cite{DBLP:conf/emnlp/WolfDSCDMCRLFDS20}.
For BERT-base and RoBERTa-base, we start from the pre-trained checkpoints of their original papers.
For BERT-large and RoBERTa-large, we utilize the checkpoints of SimCSE for stabilizing the convergence process.
Following SimCSE~\cite{DBLP:conf/emnlp/GaoYC21}, we use 1,000,000 sentences randomly sampled from Wikipedia as the training corpus.
During training, we train our models for 3 epoch with temperature $\tau=0.05$ using an Adam optimizer~\cite{DBLP:journals/corr/KingmaB14}.
For BERT-base and RoBERTa-base, the batch size is 128, the learning rate is 3e-5.
For BERT-large and RoBERTa-large, the batch size is 256, the learning rate is 3e-5 and 1e-5, respectively.
For the four backbone models, we set the instance weighting threshold $\phi$ as 0.9, 0.85, 0.9 and 0.85, respectively.
For each batch, we generate $k\times batch\_size$ noise-based negatives as the shared negatives of all instance within it, and $k$ is 1, 2.5, 4 and 5 for BERT-base, RoBERTa-base, BERT-large and RoBERTa-large, respectively.
The standard variance of the noise-based negatives is 1, and we update the noise-based negatives four times with the learning rate of 1e-3.
We evaluate the model every 150 steps on the development set of STS-B and SICK-R and keep the best checkpoint for evaluation on test sets.

\subsection{Main Results}
To verify the effectiveness of our framework on PLMs, we select BERT-base and RoBERTa-base as the base model.
Table~\ref{tab:main_sts} shows the results of different methods on seven STS tasks.

Based on the results, we can find that the non-BERT methods (\ie GloVe and USE) mostly outperform native PLM representation based baselines (\ie CLS, Mean and First-Last AVG). The reason is that directly utilizing the PLM native representations is prone to be influenced by the anisotropy issue. 
Among non-BERT methods, USE outperforms Glove. A potential reason is that USE encodes the sentence using the Transformer model, which is more effective than simply averaging GloVe embeddings. 

For other PLM-based approaches, first, we can see that flow and whitening achieve similar results and outperform the native representations based methods by a margin. These two methods adopt specific improvement strategies to refine the representations of PLMs.
Second, approaches based on contrastive learning outperform the other baselines in most cases. Contrastive learning can enhance both the alignment between semantically related positive pairs and the uniformity of the representation space using negative samples, resulting in better sentence representations.
Furthermore, SimCSE performs the best among all the baselines. It indicates that dropout is a more effective positive augmentation method than others since it rarely hurts the semantics of the sentence.

Finally, DCLR performs better than all the baselines in most settings, including the approaches based on contrastive learning. 
Since these methods mostly utilize randomly sampled negatives (\eg in-batch negatives) to learn the uniformity of all sentence representations, it may lead to sampling bias, such as false negatives and anisotropy representations.
Different from these methods, our framework adopts an instance weighting method to punish false negatives and a gradient-based algorithm to generate noise-based negatives towards the nonuniform points.
In this way, the sampling bias problem can be alleviated, and our model can better learn the uniformity to improve the quality of the sentence representations.

\section{Experiment - Analysis and Extension}
In this section, we continue to study the effectiveness of our proposed DCLR.

\begin{table}[t]
\begin{center}
\centering
\small
\begin{tabular}{l|c}
\hline
\textbf{Model} & \textbf{STS-Avg.} \\
\hline
BERT-base+Ours& \textbf{77.22} \\
\hline
w/o Noise-based Negatives & 76.17 \\
w/o Instance Weighting & 76.31 \\
\hline
BERT-base+Random Noise  & 75.22 \\
BERT-base+Knowledge Distillation & 75.05 \\
BERT-base+Self Instance Weighting & 73.93 \\
\hline
\end{tabular}
\end{center}

\caption{
Ablation and variation studies of our approach on the test set of seven STS tasks.
}
\label{tab:ablation}
\end{table}

\subsection{Debiased Contrastive Learning on Other Methods}
\label{sec:others}
\begin{figure}[t]
\centering
\includegraphics[width=0.46\textwidth]{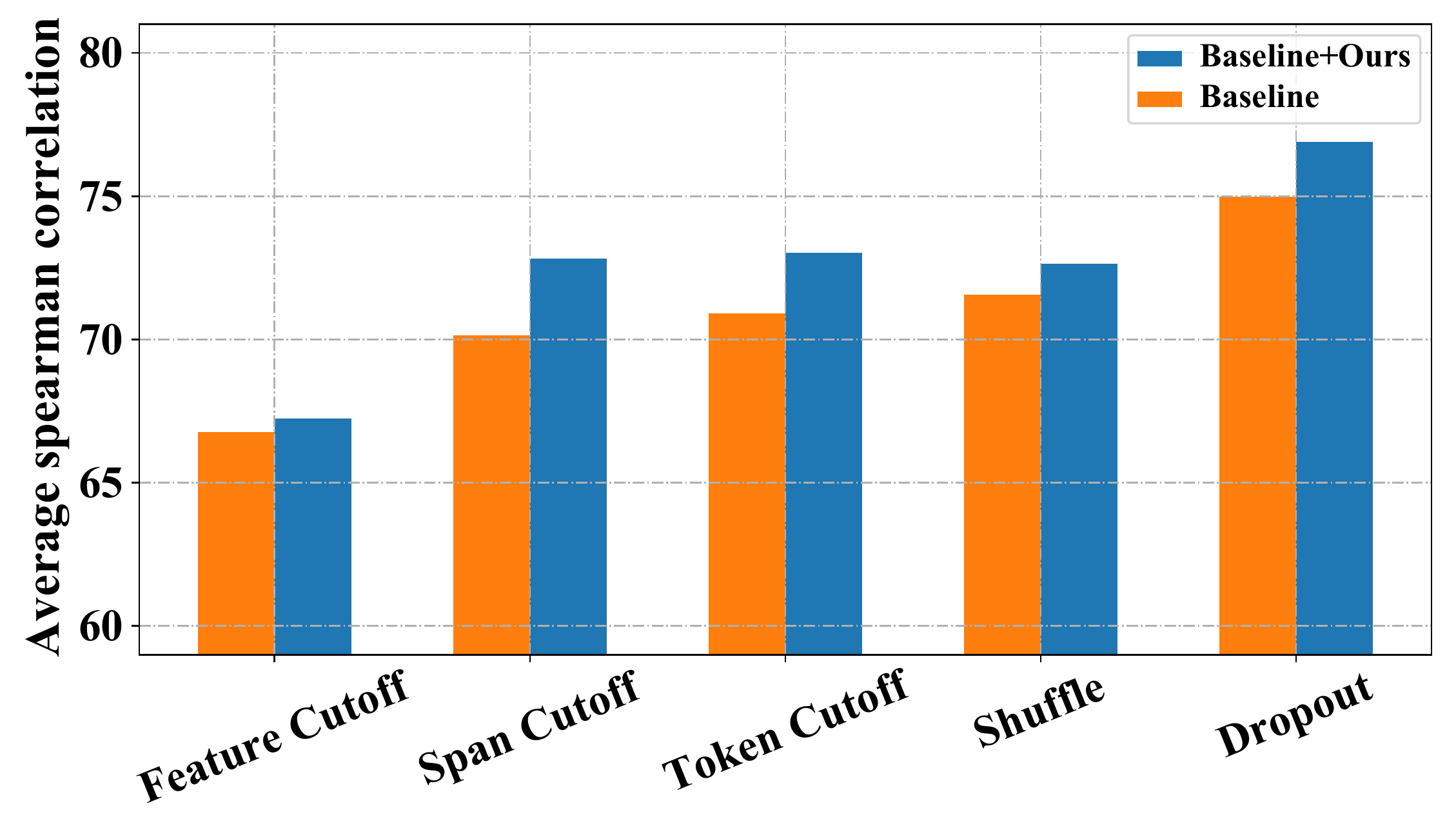}
\caption{Performance comparison using different positive augmentation strategies on the test set of seven STS tasks.}
\label{strategy}
\end{figure}

Since our proposed DCLR is a general framework that mainly focuses on negative sampling for contrastive learning of unsupervised sentence representations, it can be applied to other methods that rely on different positive data augmentation strategies. Thus, in this part, we conduct experiments to examine whether our framework can bring improvements with the following positive data augmentation strategies: (1) \emph{Token Shuffling} that randomly shuffles the order of the tokens in the input sequences; (2) \emph{Feature/Token/Span Cutoff}~\cite{DBLP:conf/acl/YanLWZWX20} that randomly erases features/tokens/token spans in the input; (3) \emph{Dropout} that is similar to SimCSE~\cite{DBLP:conf/emnlp/GaoYC21}.
Note that we only revise the negative sampling strategies to implement these variants of our DCLR.

As shown in Figure~{\ref{strategy}}, our DCLR can boost the performance of all these augmentation strategies, it demonstrates the effectiveness of our framework with various augmentation strategies. Furthermore, the Dropout strategy leads to the best performance among all the variants.  It indicates that dropout is a more effective approach to augment high-quality positives, and is also more appropriate for our approach.

\subsection{Ablation Study}
Our proposed DCLR incorporates an instance weighting method to punish false negatives and also utilizes noise-based negatives to improve the uniformity of the whole sentence representation space. To verify their effectiveness, we conduct an ablation study for each of the two components on seven STS tasks and report the average value of the Spearman's correlation metric. As shown in Table~\ref{tab:ablation}, removing each component would lead to the performance degradation. It indicates that the instance weighting method and the noise-based negatives are both important in our framework. Besides, removing the instance weighting method results in a larger performance drop. The reason may be that the false negatives have a larger effect on sentence representation learning.

Besides, we prepare three variants for further comparison:  (1) \emph{Random Noise} directly generates noise-based negatives without the gradient-based optimization; (2) \emph{Knowledge Distillation}~\cite{DBLP:journals/corr/HintonVD15} utilizes SimCSE as the teacher model to distill knowledge into the student model during training; (3) \emph{Self Instance Weighting} adopts the model itself as the complementary model to generate the weights. From Table~\ref{tab:ablation}, we can see that these variations don't perform as well as the original DCLR. These results indicate the proposed designs in Section 4 are more suitable for our DCLR framework. 

\subsection{Uniformity Analysis}
\begin{figure}[t]
\centering
\includegraphics[width=0.46\textwidth]{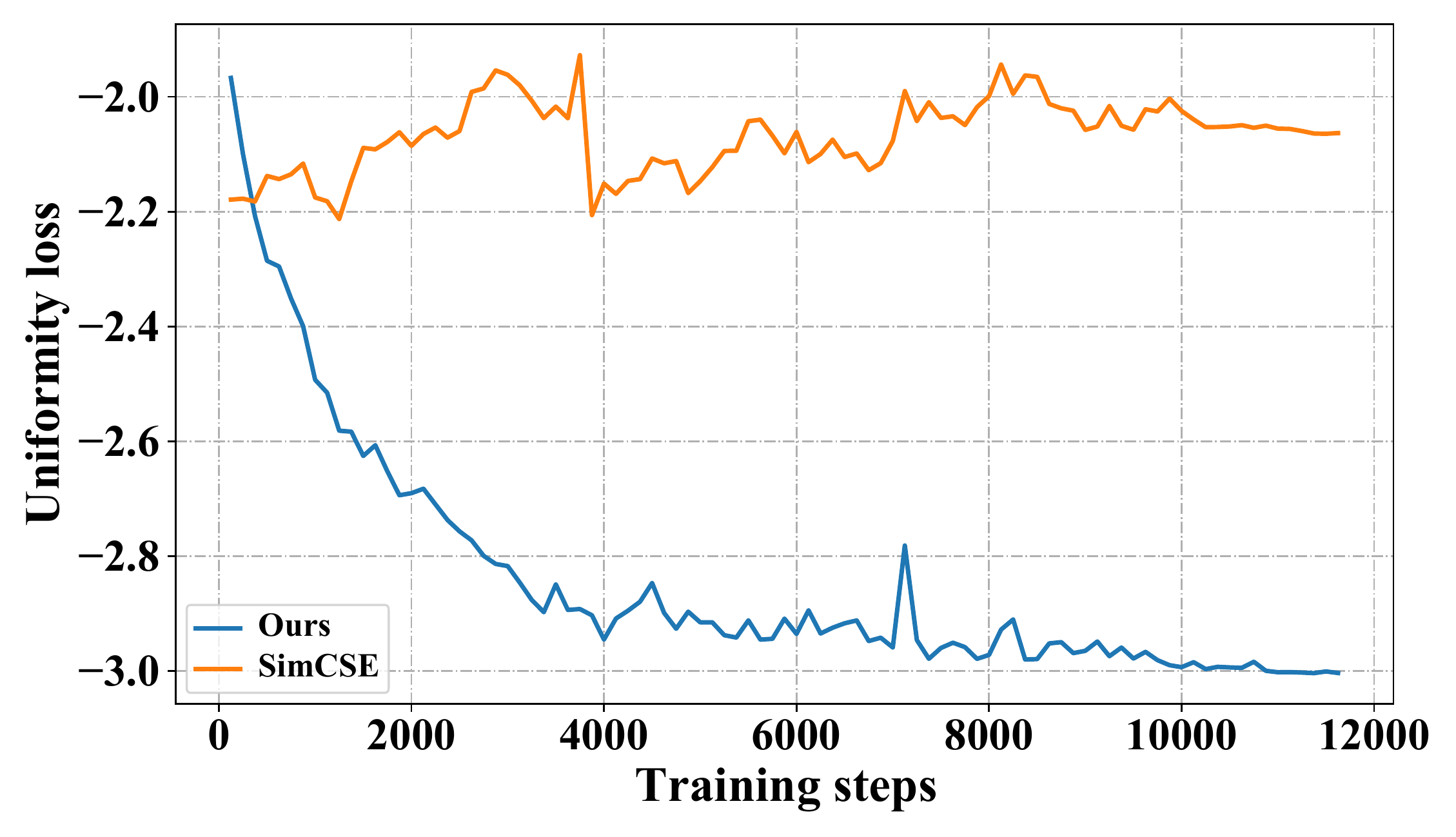}
\caption{The uniformity loss of DCLR and SimCSE using BERT-base on the validation set of STS-B during training.}
\label{uniformity}
\end{figure}

Uniformity is a desirable characteristic for sentence representations, describing how well the representations are uniformly distributed.
To validate the improvement of the uniformity of our framework, we compare the uniformity loss curves of DCLR and SimCSE using BERT-base during training. 

Following SimCSE~\cite{DBLP:conf/emnlp/GaoYC21}, we utilize the following function to evaluate the uniformity:
\begin{equation}
\label{eq:uniformity}
\ell_{uniform}\triangleq\log \underset{~~~x_i, x_j\stackrel{i.i.d.}{\sim} p_{data}}{\mathbb{E}}   e^{-2\Vert f(x_i)-f(x_j) \Vert^2}, \nonumber
\end{equation}
where $p_{data}$ is the distribution of all sentence representations, and a smaller value of this loss indicates a better uniformity.
As shown in Figure~\ref{uniformity}, the uniformity loss of DCLR is much lower than that of SimCSE in almost the whole training process. Furthermore, we can see that the uniformity loss of DCLR decreases faster as training goes, while the one of SimCSE shows no significant decreasing trend. It might be because our DCLR samples noise-based negatives beyond the representation space, which can better improve the uniformity of sentence representations.

\subsection{Performance under Few-shot Settings}

\begin{figure}[t]
\centering
\includegraphics[width=0.46\textwidth]{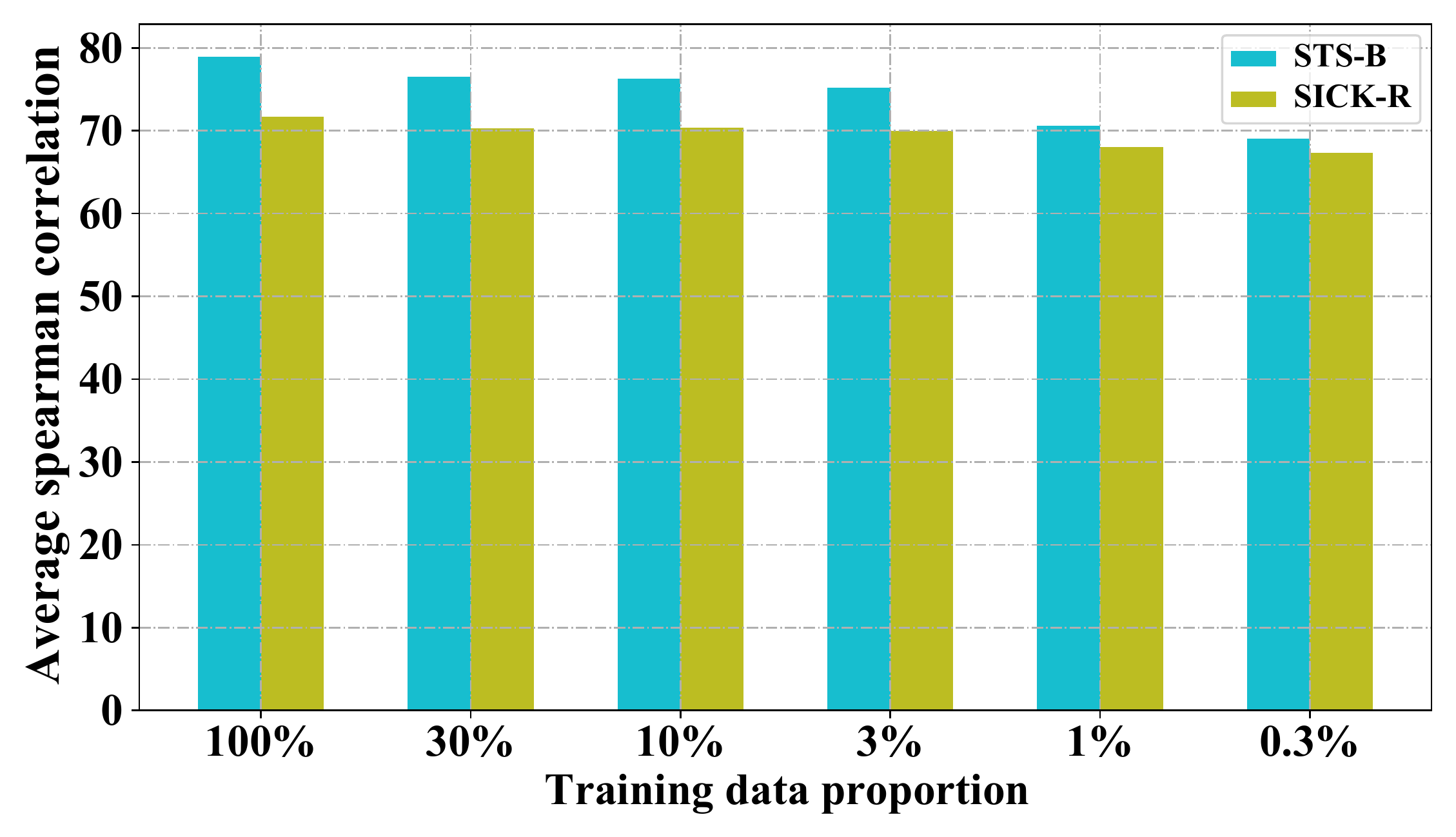}
\caption{Performance tuning of our DCLR \emph{w.r.t.} different amounts of training data. }
\label{fewshot}
\end{figure}

To validate the reliability and the robustness of DCLR under the data scarcity scenarios, we conduct few-shot experiments using BERT-base as the backbone model. 
We train our model via different amounts of available training data from 100\% to the extremely small size (\ie 0.3\%). We report the results evaluated on STS-B and SICK-R tasks.

As shown in Figure~\ref{fewshot}, our approach achieves stable results under different proportions of the training data.
Under the most extreme setting with 0.3\% data proportion, the performance of our model drops by only 9 and 4 percent on STS-B and SICK-R, respectively.
The results reveal the robustness and effectiveness of our approach under the data scarcity scenarios.
Such characteristics are important in real-world application.

\subsection{Hyper-parameters Analysis}
\begin{figure}[t!]
    \centering
    \begin{subfigure}[b]{0.49\linewidth}
        \centering
        \includegraphics[width=\textwidth]{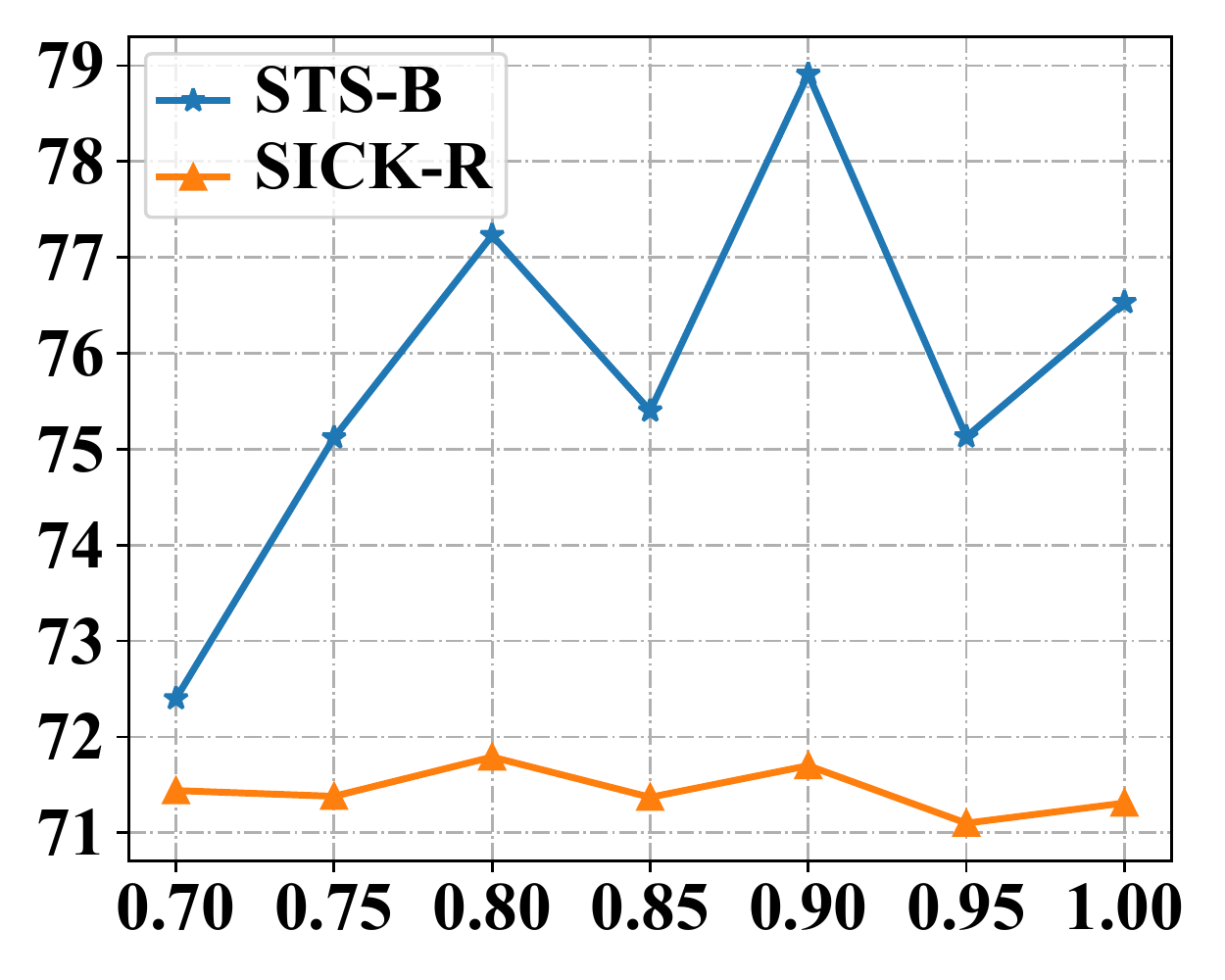}
        \caption{Weighting Threshold $\phi$}
        \label{Weighting_Threshold}
    \end{subfigure}
    \begin{subfigure}[b]{0.49\linewidth}
        \centering
        \includegraphics[width=\textwidth]{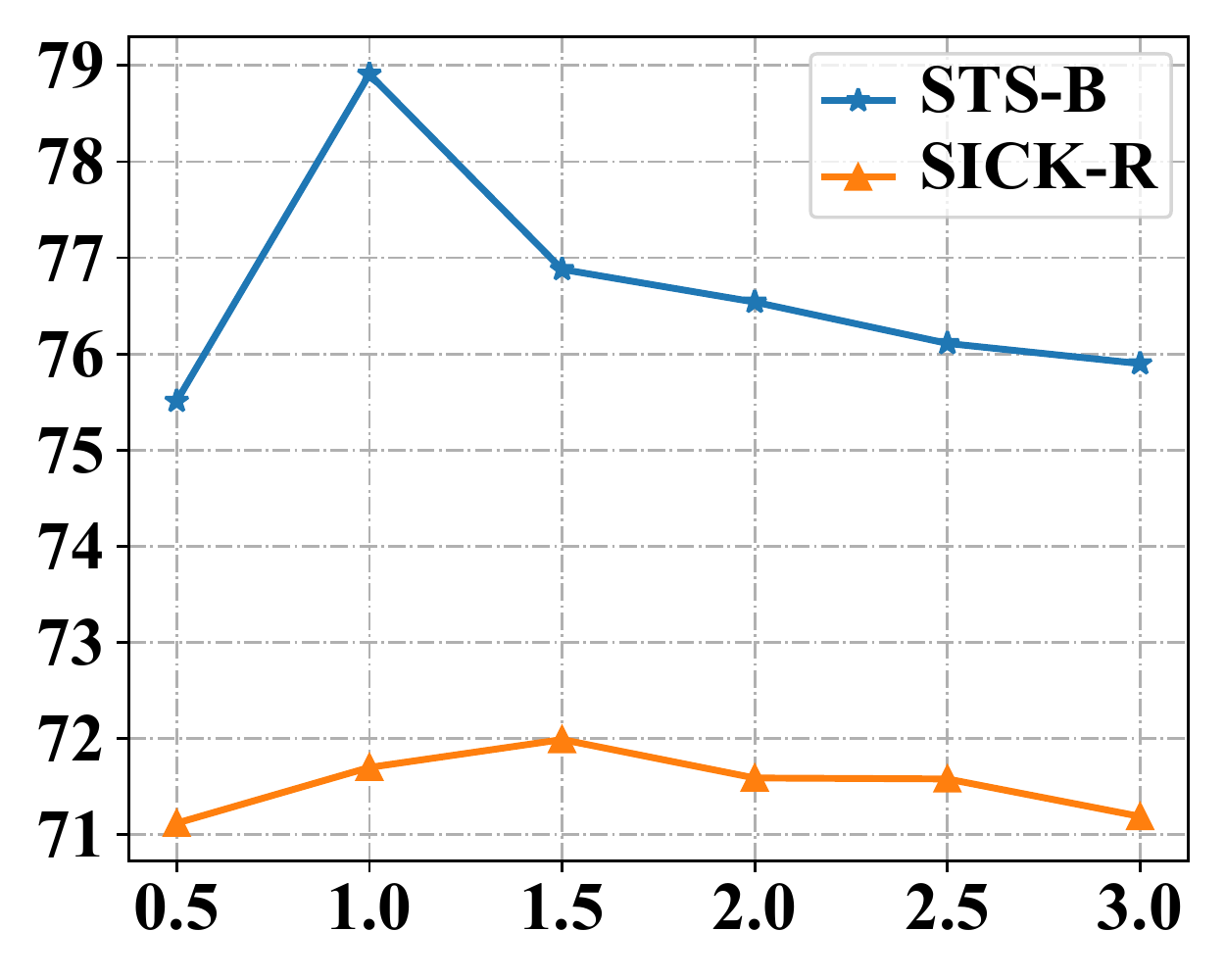}
        \caption{Negative Proportion $k$}
        \label{Negative_Proportion}
    \end{subfigure}
    \caption{Performance tuning \emph{w.r.t.} $\phi$ and $k$.}
    \vspace{-0.2cm}
\label{fig-data-amount}
\end{figure}
For hyper-parameters analysis, we study the impact of instance weighting threshold $\phi$ and the proportion of noise-based negatives $k$. 
The $\phi$ is the threshold to punish false negatives, and $k$ is the ratio of the noise-based negatives to the batch size.
Both hyper-parameters are important in our framework. 
Concretely, we evaluate our model with varying values of $\phi$ and $k$ on the STS-B and SICK-R tasks using the BERT-base model. 

\paragraph{Weighting threshold.} Figure~\ref{fig-data-amount}(a) shows the influence of the instance weighting threshold $\phi$. 
For the STS-B tasks, $\phi$ has a significant effect on the model performance. Too large or too small $\phi$ may lead to a performance drop. The reason is that a larger threshold cannot achieve effective punishment and a smaller one may cause misjudgment of true negatives. In contrast, the SICK-R is insensitive to the changes of $\phi$. The reason may be that the problem of false negatives is not serious in this task.

\paragraph{Negative proportion.} As shown in Figure~\ref{fig-data-amount}(b), our DCLR performs better when the number of noise-based negatives is close to the batch size. 
Under these circumstances, the noise-based negatives are more capable to enhance the uniformity of the whole semantic space without hurting the alignment, which is key why DCLR works well.

\section{Conclusion}
In this paper, we proposed DCLR, a debiased contrastive learning framework for unsupervised sentence representation learning. 
Our core idea is to alleviate the sampling bias caused by the random negative sampling strategy.
To achieve it, in our framework, we incorporated an instance weighting method to punish false negatives during training and generated noise-based negatives to alleviate the influence of anisotropy PLM-derived representation.
Experimental results on seven STS tasks have shown that our approach outperforms several competitive baselines.

In the future, we will explore other approaches to reducing the bias in contrastive learning of sentence representations (\eg debiased pre-training). Besides, we will also consider to apply our method for multilingual or multimodal representation learning.

\section*{Ethical Consideration}
In this section, we discuss the ethical considerations of this work from the following two aspects.
First, for intellectual property protection, the code, data and pre-trained models adopted from previous works are granted for research-purpose usage.
Second, since PLMs have been shown to capture certain biases from the data they have been pre-trained on~\cite{bender2021dangers},
there is a potential problem about biases that are from the use of PLMs in our approach.
There are increasing efforts to address this problem in the community~\cite{ross2020measuring}.

\section*{Acknowledgement}
This work was partially supported by Beijing Natural Science Foundation under Grant No. 4222027, and  National Natural Science Foundation of China under Grant No. 61872369, Beijing Outstanding Young Scientist Program under Grant No. BJJWZYJH012019100020098, 
the Outstanding Innovative Talents Cultivation Funded Programs 2021 and Public Computing Cloud, Renmin University of China. This work is also supported by Beijing Academy of Artificial Intelligence~(BAAI). Xin Zhao is the corresponding author.

\bibliography{anthology,custom}

\begin{thebibliography}{49}
\expandafter\ifx\csname natexlab\endcsname\relax\def\natexlab#1{#1}\fi

\bibitem[{Agirre et~al.(2015)Agirre, Banea, Cardie, Cer, Diab,
  Gonzalez{-}Agirre, Guo, Lopez{-}Gazpio, Maritxalar, Mihalcea, Rigau, Uria,
  and Wiebe}]{DBLP:conf/semeval/AgirreBCCDGGLMM15}
Eneko Agirre, Carmen Banea, Claire Cardie, Daniel~M. Cer, Mona~T. Diab, Aitor
  Gonzalez{-}Agirre, Weiwei Guo, I{\~{n}}igo Lopez{-}Gazpio, Montse Maritxalar,
  Rada Mihalcea, German Rigau, Larraitz Uria, and Janyce Wiebe. 2015.
\newblock \href {https://doi.org/10.18653/v1/s15-2045} {Semeval-2015 task 2:
  Semantic textual similarity, english, spanish and pilot on interpretability}.
\newblock In \emph{{NAACL-HLT}}, pages 252--263.

\bibitem[{Agirre et~al.(2014)Agirre, Banea, Cardie, Cer, Diab,
  Gonzalez{-}Agirre, Guo, Mihalcea, Rigau, and
  Wiebe}]{DBLP:conf/semeval/AgirreBCCDGGMRW14}
Eneko Agirre, Carmen Banea, Claire Cardie, Daniel~M. Cer, Mona~T. Diab, Aitor
  Gonzalez{-}Agirre, Weiwei Guo, Rada Mihalcea, German Rigau, and Janyce Wiebe.
  2014.
\newblock \href {https://doi.org/10.3115/v1/s14-2010} {Semeval-2014 task 10:
  Multilingual semantic textual similarity}.
\newblock In \emph{{COLING}}, pages 81--91.

\bibitem[{Agirre et~al.(2016)Agirre, Banea, Cer, Diab, Gonzalez{-}Agirre,
  Mihalcea, Rigau, and Wiebe}]{DBLP:conf/semeval/AgirreBCDGMRW16}
Eneko Agirre, Carmen Banea, Daniel~M. Cer, Mona~T. Diab, Aitor
  Gonzalez{-}Agirre, Rada Mihalcea, German Rigau, and Janyce Wiebe. 2016.
\newblock \href {https://doi.org/10.18653/v1/s16-1081} {Semeval-2016 task 1:
  Semantic textual similarity, monolingual and cross-lingual evaluation}.
\newblock In \emph{{COLING}}, pages 497--511.

\bibitem[{Agirre et~al.(2012)Agirre, Cer, Diab, and
  Gonzalez{-}Agirre}]{DBLP:conf/semeval/AgirreCDG12}
Eneko Agirre, Daniel~M. Cer, Mona~T. Diab, and Aitor Gonzalez{-}Agirre. 2012.
\newblock \href {https://aclanthology.org/S12-1051/} {Semeval-2012 task 6: {A}
  pilot on semantic textual similarity}.
\newblock In \emph{{NAACL-HLT}}, pages 385--393.

\bibitem[{Agirre et~al.(2013)Agirre, Cer, Diab, Gonzalez{-}Agirre, and
  Guo}]{DBLP:conf/starsem/AgirreCDGG13}
Eneko Agirre, Daniel~M. Cer, Mona~T. Diab, Aitor Gonzalez{-}Agirre, and Weiwei
  Guo. 2013.
\newblock \href {https://aclanthology.org/S13-1004/} {*sem 2013 shared task:
  Semantic textual similarity}.
\newblock In \emph{*SEM}, pages 32--43.

\bibitem[{Bender et~al.(2021)Bender, Gebru, McMillan-Major, and
  Shmitchell}]{bender2021dangers}
Emily~M Bender, Timnit Gebru, Angelina McMillan-Major, and Shmargaret
  Shmitchell. 2021.
\newblock On the dangers of stochastic parrots: Can language models be too big?
\newblock In \emph{Proceedings of the 2021 ACM Conference on Fairness,
  Accountability, and Transparency}, pages 610--623.

\bibitem[{Bian et~al.(2021)Bian, Zhao, Zhou, Cai, He, Yin, and
  Wen}]{DBLP:conf/cikm/BianZZCHYW21}
Shuqing Bian, Wayne~Xin Zhao, Kun Zhou, Jing Cai, Yancheng He, Cunxiang Yin,
  and Ji{-}Rong Wen. 2021.
\newblock \href {https://doi.org/10.1145/3459637.3481905} {Contrastive
  curriculum learning for sequential user behavior modeling via data
  augmentation}.
\newblock In \emph{{CIKM} '21: The 30th {ACM} International Conference on
  Information and Knowledge Management, Virtual Event, Queensland, Australia,
  November 1 - 5, 2021}, pages 3737--3746. {ACM}.

\bibitem[{Bowman et~al.(2015)Bowman, Angeli, Potts, and
  Manning}]{DBLP:conf/emnlp/BowmanAPM15}
Samuel~R. Bowman, Gabor Angeli, Christopher Potts, and Christopher~D. Manning.
  2015.
\newblock \href {https://doi.org/10.18653/v1/d15-1075} {A large annotated
  corpus for learning natural language inference}.
\newblock In \emph{{EMNLP}}, pages 632--642.

\bibitem[{Cer et~al.(2018)Cer, Yang, Kong, Hua, Limtiaco, John, Constant,
  Guajardo{-}Cespedes, Yuan, Tar, Strope, and
  Kurzweil}]{DBLP:conf/emnlp/CerYKHLJCGYTSK18}
Daniel Cer, Yinfei Yang, Sheng{-}yi Kong, Nan Hua, Nicole Limtiaco, Rhomni~St.
  John, Noah Constant, Mario Guajardo{-}Cespedes, Steve Yuan, Chris Tar, Brian
  Strope, and Ray Kurzweil. 2018.
\newblock \href {https://doi.org/10.18653/v1/d18-2029} {Universal sentence
  encoder for english}.
\newblock In \emph{{EMNLP}}, pages 169--174.

\bibitem[{Cer et~al.(2017)Cer, Diab, Agirre, Lopez{-}Gazpio, and
  Specia}]{DBLP:conf/semeval/CerDALS17}
Daniel~M. Cer, Mona~T. Diab, Eneko Agirre, I{\~{n}}igo Lopez{-}Gazpio, and
  Lucia Specia. 2017.
\newblock \href {https://doi.org/10.18653/v1/S17-2001} {Semeval-2017 task 1:
  Semantic textual similarity multilingual and crosslingual focused
  evaluation}.
\newblock In \emph{ACL}, pages 1--14.

\bibitem[{Chen et~al.(2020)Chen, Kornblith, Norouzi, and
  Hinton}]{DBLP:conf/icml/ChenK0H20}
Ting Chen, Simon Kornblith, Mohammad Norouzi, and Geoffrey~E. Hinton. 2020.
\newblock \href {http://proceedings.mlr.press/v119/chen20j.html} {A simple
  framework for contrastive learning of visual representations}.
\newblock In \emph{{ICML}}, volume 119 of \emph{Proceedings of Machine Learning
  Research}, pages 1597--1607.

\bibitem[{Conneau and Kiela(2018)}]{DBLP:conf/lrec/ConneauK18}
Alexis Conneau and Douwe Kiela. 2018.
\newblock \href
  {http://www.lrec-conf.org/proceedings/lrec2018/summaries/757.html} {Senteval:
  An evaluation toolkit for universal sentence representations}.
\newblock In \emph{{LREC}}.

\bibitem[{Conneau et~al.(2017)Conneau, Kiela, Schwenk, Barrault, and
  Bordes}]{DBLP:conf/emnlp/ConneauKSBB17}
Alexis Conneau, Douwe Kiela, Holger Schwenk, Lo{\"{\i}}c Barrault, and Antoine
  Bordes. 2017.
\newblock \href {https://doi.org/10.18653/v1/d17-1070} {Supervised learning of
  universal sentence representations from natural language inference data}.
\newblock In \emph{{EMNLP}}, pages 670--680.

\bibitem[{Devlin et~al.(2019)Devlin, Chang, Lee, and
  Toutanova}]{DBLP:conf/naacl/DevlinCLT19}
Jacob Devlin, Ming{-}Wei Chang, Kenton Lee, and Kristina Toutanova. 2019.
\newblock \href {https://doi.org/10.18653/v1/n19-1423} {{BERT:} pre-training of
  deep bidirectional transformers for language understanding}.
\newblock In \emph{{NAACL-HLT}}, pages 4171--4186.

\bibitem[{Ethayarajh(2019)}]{DBLP:conf/emnlp/Ethayarajh19}
Kawin Ethayarajh. 2019.
\newblock \href {https://doi.org/10.18653/v1/D19-1006} {How contextual are
  contextualized word representations? comparing the geometry of bert, elmo,
  and {GPT-2} embeddings}.
\newblock In \emph{{EMNLP-IJCNLP}}, pages 55--65.

\bibitem[{Fang and Xie(2020)}]{DBLP:journals/corr/abs-2005-12766}
Hongchao Fang and Pengtao Xie. 2020.
\newblock \href {http://arxiv.org/abs/2005.12766} {{CERT:} contrastive
  self-supervised learning for language understanding}.
\newblock \emph{CoRR}, abs/2005.12766.

\bibitem[{Gao et~al.(2021)Gao, Yao, and Chen}]{DBLP:conf/emnlp/GaoYC21}
Tianyu Gao, Xingcheng Yao, and Danqi Chen. 2021.
\newblock \href {https://aclanthology.org/2021.emnlp-main.552} {Simcse: Simple
  contrastive learning of sentence embeddings}.
\newblock In \emph{{EMNLP}}, pages 6894--6910. Association for Computational
  Linguistics.

\bibitem[{Hadsell et~al.(2006)Hadsell, Chopra, and
  LeCun}]{DBLP:conf/cvpr/HadsellCL06}
Raia Hadsell, Sumit Chopra, and Yann LeCun. 2006.
\newblock \href {https://doi.org/10.1109/CVPR.2006.100} {Dimensionality
  reduction by learning an invariant mapping}.
\newblock In \emph{{CVPR}}, pages 1735--1742.

\bibitem[{He et~al.(2020)He, Fan, Wu, Xie, and
  Girshick}]{DBLP:conf/cvpr/He0WXG20}
Kaiming He, Haoqi Fan, Yuxin Wu, Saining Xie, and Ross~B. Girshick. 2020.
\newblock \href {https://doi.org/10.1109/CVPR42600.2020.00975} {Momentum
  contrast for unsupervised visual representation learning}.
\newblock In \emph{{CVPR}}, pages 9726--9735.

\bibitem[{Hill et~al.(2016)Hill, Cho, and Korhonen}]{DBLP:conf/naacl/HillCK16}
Felix Hill, Kyunghyun Cho, and Anna Korhonen. 2016.
\newblock \href {https://doi.org/10.18653/v1/n16-1162} {Learning distributed
  representations of sentences from unlabelled data}.
\newblock In \emph{{NAACL-HLT}}, pages 1367--1377.

\bibitem[{Hinton et~al.(2015)Hinton, Vinyals, and
  Dean}]{DBLP:journals/corr/HintonVD15}
Geoffrey~E. Hinton, Oriol Vinyals, and Jeffrey Dean. 2015.
\newblock \href {http://arxiv.org/abs/1503.02531} {Distilling the knowledge in
  a neural network}.
\newblock \emph{CoRR}, abs/1503.02531.

\bibitem[{Huang et~al.(2021)Huang, Tang, Zhong, Lu, Shou, Gong, Jiang, and
  Duan}]{DBLP:journals/corr/abs-2104-01767}
Junjie Huang, Duyu Tang, Wanjun Zhong, Shuai Lu, Linjun Shou, Ming Gong, Daxin
  Jiang, and Nan Duan. 2021.
\newblock \href {http://arxiv.org/abs/2104.01767} {Whiteningbert: An easy
  unsupervised sentence embedding approach}.
\newblock \emph{CoRR}, abs/2104.01767.

\bibitem[{Jiang et~al.(2020)Jiang, He, Chen, Liu, Gao, and
  Zhao}]{DBLP:conf/acl/JiangHCLGZ20}
Haoming Jiang, Pengcheng He, Weizhu Chen, Xiaodong Liu, Jianfeng Gao, and Tuo
  Zhao. 2020.
\newblock \href {https://doi.org/10.18653/v1/2020.acl-main.197} {{SMART:}
  robust and efficient fine-tuning for pre-trained natural language models
  through principled regularized optimization}.
\newblock In \emph{{ACL}}, pages 2177--2190.

\bibitem[{Kim et~al.(2021)Kim, Yoo, and Lee}]{DBLP:conf/acl/KimYL20}
Taeuk Kim, Kang~Min Yoo, and Sang{-}goo Lee. 2021.
\newblock \href {https://doi.org/10.18653/v1/2021.acl-long.197} {Self-guided
  contrastive learning for {BERT} sentence representations}.
\newblock In \emph{{ACL}}, pages 2528--2540.

\bibitem[{Kingma and Ba(2015)}]{DBLP:journals/corr/KingmaB14}
Diederik~P. Kingma and Jimmy Ba. 2015.
\newblock \href {http://arxiv.org/abs/1412.6980} {Adam: {A} method for
  stochastic optimization}.
\newblock In \emph{{ICLR}}.

\bibitem[{Kiros et~al.(2015)Kiros, Zhu, Salakhutdinov, Zemel, Urtasun,
  Torralba, and Fidler}]{DBLP:conf/nips/KirosZSZUTF15}
Ryan Kiros, Yukun Zhu, Ruslan Salakhutdinov, Richard~S. Zemel, Raquel Urtasun,
  Antonio Torralba, and Sanja Fidler. 2015.
\newblock \href
  {https://proceedings.neurips.cc/paper/2015/hash/f442d33fa06832082290ad8544a8da27-Abstract.html}
  {Skip-thought vectors}.
\newblock In \emph{Advances in Neural Information Processing Systems 28: Annual
  Conference on Neural Information Processing Systems 2015, December 7-12,
  2015, Montreal, Quebec, Canada}, pages 3294--3302.

\bibitem[{Kurakin et~al.(2017)Kurakin, Goodfellow, and
  Bengio}]{DBLP:conf/iclr/KurakinGB17a}
Alexey Kurakin, Ian~J. Goodfellow, and Samy Bengio. 2017.
\newblock \href {https://openreview.net/forum?id=HJGU3Rodl} {Adversarial
  examples in the physical world}.
\newblock In \emph{{ICLR}}.

\bibitem[{Le and Mikolov(2014)}]{DBLP:conf/icml/LeM14}
Quoc~V. Le and Tom{\'{a}}s Mikolov. 2014.
\newblock \href {http://proceedings.mlr.press/v32/le14.html} {Distributed
  representations of sentences and documents}.
\newblock In \emph{{ICML}}, volume~32 of \emph{{JMLR} Workshop and Conference
  Proceedings}, pages 1188--1196.

\bibitem[{Li et~al.(2020)Li, Zhou, He, Wang, Yang, and
  Li}]{DBLP:conf/emnlp/LiZHWYL20}
Bohan Li, Hao Zhou, Junxian He, Mingxuan Wang, Yiming Yang, and Lei Li. 2020.
\newblock \href {https://doi.org/10.18653/v1/2020.emnlp-main.733} {On the
  sentence embeddings from pre-trained language models}.
\newblock In \emph{{EMNLP}}, pages 9119--9130.

\bibitem[{Liu et~al.(2019)Liu, Ott, Goyal, Du, Joshi, Chen, Levy, Lewis,
  Zettlemoyer, and Stoyanov}]{DBLP:journals/corr/abs-1907-11692}
Yinhan Liu, Myle Ott, Naman Goyal, Jingfei Du, Mandar Joshi, Danqi Chen, Omer
  Levy, Mike Lewis, Luke Zettlemoyer, and Veselin Stoyanov. 2019.
\newblock \href {http://arxiv.org/abs/1907.11692} {Roberta: {A} robustly
  optimized {BERT} pretraining approach}.
\newblock \emph{CoRR}, abs/1907.11692.

\bibitem[{Madry et~al.(2018)Madry, Makelov, Schmidt, Tsipras, and
  Vladu}]{DBLP:conf/iclr/MadryMSTV18}
Aleksander Madry, Aleksandar Makelov, Ludwig Schmidt, Dimitris Tsipras, and
  Adrian Vladu. 2018.
\newblock \href {https://openreview.net/forum?id=rJzIBfZAb} {Towards deep
  learning models resistant to adversarial attacks}.
\newblock In \emph{{ICLR}}.

\bibitem[{Marelli et~al.(2014)Marelli, Menini, Baroni, Bentivogli, Bernardi,
  and Zamparelli}]{DBLP:conf/lrec/MarelliMBBBZ14}
Marco Marelli, Stefano Menini, Marco Baroni, Luisa Bentivogli, Raffaella
  Bernardi, and Roberto Zamparelli. 2014.
\newblock \href
  {http://www.lrec-conf.org/proceedings/lrec2014/summaries/363.html} {A {SICK}
  cure for the evaluation of compositional distributional semantic models}.
\newblock In \emph{{LREC}}, pages 216--223.

\bibitem[{Mikolov et~al.(2013)Mikolov, Sutskever, Chen, Corrado, and
  Dean}]{DBLP:conf/nips/MikolovSCCD13}
Tom{\'{a}}s Mikolov, Ilya Sutskever, Kai Chen, Gregory~S. Corrado, and Jeffrey
  Dean. 2013.
\newblock \href
  {https://proceedings.neurips.cc/paper/2013/hash/9aa42b31882ec039965f3c4923ce901b-Abstract.html}
  {Distributed representations of words and phrases and their
  compositionality}.
\newblock In \emph{Advances in Neural Information Processing Systems 26: 27th
  Annual Conference on Neural Information Processing Systems 2013. Proceedings
  of a meeting held December 5-8, 2013, Lake Tahoe, Nevada, United States},
  pages 3111--3119.

\bibitem[{Miyato et~al.(2017)Miyato, Dai, and
  Goodfellow}]{DBLP:conf/iclr/MiyatoDG17}
Takeru Miyato, Andrew~M. Dai, and Ian~J. Goodfellow. 2017.
\newblock \href {https://openreview.net/forum?id=r1X3g2\_xl} {Adversarial
  training methods for semi-supervised text classification}.
\newblock In \emph{{ICLR}}.

\bibitem[{Miyato et~al.(2019)Miyato, Maeda, Koyama, and
  Ishii}]{DBLP:journals/pami/MiyatoMKI19}
Takeru Miyato, Shin{-}ichi Maeda, Masanori Koyama, and Shin Ishii. 2019.
\newblock \href {https://doi.org/10.1109/TPAMI.2018.2858821} {Virtual
  adversarial training: {A} regularization method for supervised and
  semi-supervised learning}.
\newblock \emph{{IEEE} Trans. Pattern Anal. Mach. Intell.}, 41(8):1979--1993.

\bibitem[{Pennington et~al.(2014)Pennington, Socher, and
  Manning}]{DBLP:conf/emnlp/PenningtonSM14}
Jeffrey Pennington, Richard Socher, and Christopher~D. Manning. 2014.
\newblock \href {https://doi.org/10.3115/v1/d14-1162} {Glove: Global vectors
  for word representation}.
\newblock In \emph{{EMNLP}}, pages 1532--1543.

\bibitem[{Qiao et~al.(2016)Qiao, Liu, Shen, and Van Den~Hengel}]{qiao2016less}
Ruizhi Qiao, Lingqiao Liu, Chunhua Shen, and Anton Van Den~Hengel. 2016.
\newblock Less is more: zero-shot learning from online textual documents with
  noise suppression.
\newblock In \emph{{CVPR}}, pages 2249--2257.

\bibitem[{Qin et~al.(2019)Qin, Martens, Gowal, Krishnan, Dvijotham, Fawzi, De,
  Stanforth, and Kohli}]{DBLP:conf/nips/QinMGKDFDSK19}
Chongli Qin, James Martens, Sven Gowal, Dilip Krishnan, Krishnamurthy
  Dvijotham, Alhussein Fawzi, Soham De, Robert Stanforth, and Pushmeet Kohli.
  2019.
\newblock \href
  {https://proceedings.neurips.cc/paper/2019/hash/0defd533d51ed0a10c5c9dbf93ee78a5-Abstract.html}
  {Adversarial robustness through local linearization}.
\newblock In \emph{NeurIPS}, pages 13824--13833.

\bibitem[{Reimers and Gurevych(2019)}]{DBLP:conf/emnlp/ReimersG19}
Nils Reimers and Iryna Gurevych. 2019.
\newblock \href {https://doi.org/10.18653/v1/D19-1410} {Sentence-bert: Sentence
  embeddings using siamese bert-networks}.
\newblock In \emph{{EMNLP-IJCNLP}}, pages 3980--3990.

\bibitem[{Ross et~al.(2020)Ross, Katz, and Barbu}]{ross2020measuring}
Candace Ross, Boris Katz, and Andrei Barbu. 2020.
\newblock Measuring social biases in grounded vision and language embeddings.
\newblock \emph{arXiv preprint arXiv:2002.08911}.

\bibitem[{Su et~al.(2021)Su, Cao, Liu, and
  Ou}]{DBLP:journals/corr/abs-2103-15316}
Jianlin Su, Jiarun Cao, Weijie Liu, and Yangyiwen Ou. 2021.
\newblock \href {http://arxiv.org/abs/2103.15316} {Whitening sentence
  representations for better semantics and faster retrieval}.
\newblock \emph{CoRR}, abs/2103.15316.

\bibitem[{Sun et~al.(2020)Sun, Wang, Chen, Lu, Utiyama, Sumita, and
  Zhao}]{DBLP:conf/coling/SunWCLUSZ20}
Haipeng Sun, Rui Wang, Kehai Chen, Xugang Lu, Masao Utiyama, Eiichiro Sumita,
  and Tiejun Zhao. 2020.
\newblock \href {https://doi.org/10.18653/v1/2020.coling-main.374} {Robust
  unsupervised neural machine translation with adversarial denoising training}.
\newblock In \emph{{COLING}}, pages 4239--4250.

\bibitem[{Williams et~al.(2018)Williams, Nangia, and
  Bowman}]{DBLP:conf/naacl/WilliamsNB18}
Adina Williams, Nikita Nangia, and Samuel~R. Bowman. 2018.
\newblock \href {https://doi.org/10.18653/v1/n18-1101} {A broad-coverage
  challenge corpus for sentence understanding through inference}.
\newblock In \emph{{NAACL-HLT}}, pages 1112--1122.

\bibitem[{Wolf et~al.(2020)Wolf, Debut, Sanh, Chaumond, Delangue, Moi, Cistac,
  Rault, Louf, Funtowicz, Davison, Shleifer, von Platen, Ma, Jernite, Plu, Xu,
  Scao, Gugger, Drame, Lhoest, and Rush}]{DBLP:conf/emnlp/WolfDSCDMCRLFDS20}
Thomas Wolf, Lysandre Debut, Victor Sanh, Julien Chaumond, Clement Delangue,
  Anthony Moi, Pierric Cistac, Tim Rault, R{\'{e}}mi Louf, Morgan Funtowicz,
  Joe Davison, Sam Shleifer, Patrick von Platen, Clara Ma, Yacine Jernite,
  Julien Plu, Canwen Xu, Teven~Le Scao, Sylvain Gugger, Mariama Drame, Quentin
  Lhoest, and Alexander~M. Rush. 2020.
\newblock \href {https://doi.org/10.18653/v1/2020.emnlp-demos.6} {Transformers:
  State-of-the-art natural language processing}.
\newblock In \emph{{EMNLP} - Demos}, pages 38--45.

\bibitem[{Wu et~al.(2021)Wu, Gao, Zang, Han, Wang, and
  Hu}]{DBLP:journals/corr/abs-2109-04321}
Xing Wu, Chaochen Gao, Liangjun Zang, Jizhong Han, Zhongyuan Wang, and Songlin
  Hu. 2021.
\newblock \href {http://arxiv.org/abs/2109.04321} {Smoothed contrastive
  learning for unsupervised sentence embedding}.
\newblock \emph{CoRR}, abs/2109.04321.

\bibitem[{Wu et~al.(2020)Wu, Wang, Gu, Khabsa, Sun, and
  Ma}]{DBLP:journals/corr/abs-2012-15466}
Zhuofeng Wu, Sinong Wang, Jiatao Gu, Madian Khabsa, Fei Sun, and Hao Ma. 2020.
\newblock \href {http://arxiv.org/abs/2012.15466} {{CLEAR:} contrastive
  learning for sentence representation}.
\newblock \emph{CoRR}, abs/2012.15466.

\bibitem[{Yan et~al.(2021)Yan, Li, Wang, Zhang, Wu, and
  Xu}]{DBLP:conf/acl/YanLWZWX20}
Yuanmeng Yan, Rumei Li, Sirui Wang, Fuzheng Zhang, Wei Wu, and Weiran Xu. 2021.
\newblock \href {https://doi.org/10.18653/v1/2021.acl-long.393} {Consert: {A}
  contrastive framework for self-supervised sentence representation transfer}.
\newblock In \emph{{ACL/IJCNLP}}, pages 5065--5075.

\bibitem[{Zhou et~al.(2022)Zhou, Zhou, Zhao, Wang, Jiang, and
  Hu}]{DBLP:conf/wsdm/ZhouZZWJ022}
Yuanhang Zhou, Kun Zhou, Wayne~Xin Zhao, Cheng Wang, Peng Jiang, and He~Hu.
  2022.
\newblock \href {https://doi.org/10.1145/3488560.3498514} {C{\({^2}\)}-crs:
  Coarse-to-fine contrastive learning for conversational recommender system}.
\newblock In \emph{{WSDM} '22: The Fifteenth {ACM} International Conference on
  Web Search and Data Mining, Virtual Event / Tempe, AZ, USA, February 21 - 25,
  2022}, pages 1488--1496. {ACM}.

\bibitem[{Zhu et~al.(2020)Zhu, Cheng, Gan, Sun, Goldstein, and
  Liu}]{DBLP:conf/iclr/ZhuCGSGL20}
Chen Zhu, Yu~Cheng, Zhe Gan, Siqi Sun, Tom Goldstein, and Jingjing Liu. 2020.
\newblock \href {https://openreview.net/forum?id=BygzbyHFvB} {Freelb: Enhanced
  adversarial training for natural language understanding}.
\newblock In \emph{{ICLR}}.

\end{thebibliography}
\bibliographystyle{acl_natbib}




\end{document}